\documentclass{article}

\PassOptionsToPackage{numbers, compress}{natbib}
\usepackage[numbers]{natbib}

\usepackage[final]{neurips_2024}

\usepackage[utf8]{inputenc} 
\usepackage[T1]{fontenc}    
\usepackage{url}            
\usepackage{booktabs}       
\usepackage{amsfonts}       
\usepackage{nicefrac}       
\usepackage{microtype}      

\usepackage{amsmath,amssymb,amsfonts}
\usepackage{array,multirow}
\usepackage{graphicx}
\usepackage{subcaption}
\usepackage{enumitem}
\usepackage[disable]{todonotes} 
\usepackage{bbm}
\usepackage{hyperref}       

\title{Understanding The Effect Of Temperature On Alignment With Human
Opinions } 

\author{%
  Maja Pavlovic \\
  Queen Mary University of London \\
  London, United Kingdom \\
  \texttt{m.pavlovic@qmul.ac.uk} \\
  \And
  Massimo Poesio \\
  Queen Mary University of London - University of Utrecht \\
  London, United Kingdom - Utrecht, Netherlands  \\
  \texttt{m.poesio@\{qmul.ac.uk,uu.nl\}} \\
}

\begin{document}

\maketitle

\begin{abstract}
With the increasing capabilities of LLMs, recent studies focus on 
understanding whose opinions are represented by them and how to effectively extract aligned opinion distributions.
We conducted an empirical analysis of three straightforward methods for obtaining distributions and evaluated the results across a variety of metrics. 
\textcolor{black}{Our findings suggest that sampling and log-probability approaches with simple parameter adjustments can return better aligned outputs in subjective tasks compared to direct prompting.} 
Yet, assuming models reflect human opinions may be limiting, highlighting the need for further research on how human subjectivity affects model uncertainty.
\end{abstract}

\vspace{-0.2cm}
\section{Introduction}
\vspace{-0.2cm}
Large Language Models (LLMs) have shown notable skills across a diverse range of tasks. Their increasing prevalence underscores the need for effectively modeling and representing a diverse range of human values and perspectives \citep{sorensen2024roadmappluralisticalignment}.
Several studies investigated the extent to which LLMs represent multiple human opinions 
\citep{feng2024modularpluralismpluralisticalignment, chen2024seeingbigsmallllms, durmus2024measuringrepresentationsubjectiveglobal, santurkar2023whose, lee2023can, pavl2024effectiveness}.
While studies \citep{durmus2024measuringrepresentationsubjectiveglobal} and \citep{santurkar2023whose} explored whose opinions are reflected by LLMs on a subjective opinion QA task, study 
\citep{pavl2024effectiveness} assessed
their capability to capture human opinion distributions in
subjective classification tasks. In \citep{lee2023can} and \citep{chen2024seeingbigsmallllms} on the other hand, 
the authors analysed whether LLMs align with human opinion distributions in NLI {\small(natural language inference)} tasks.
Recent studies observed misaligned LLM and human opinion distributions \citep{santurkar2023whose, lee2023can, pavl2024effectiveness}. Others found methods that improve the alignment with human distributions 
\citep{chen2024seeingbigsmallllms, feng2024modularpluralismpluralisticalignment}.
However, none of the studies examined the effect of temperature. We consider that allowing models to chose beyond the maximum likelihood token {\small(0 temperature)} may reveal more diverse opinions that could align with broader human views.

\section{Related work} 
\vspace{-0.2cm}
We approach measuring human opinion alignment through the lens of distributional pluralism
\citep{sorensen2024roadmappluralisticalignment}. In study \citep{feng2024modularpluralismpluralisticalignment}, more pluralistic responses were obtained by incorporating specialised community LMs, while in \citep{chen2024seeingbigsmallllms} a small number of expert explanations were used to extract more aligned opinion distributions. Other studies adopted simpler approaches, extracting distributions without human explanations or additional models.
In \citep{pavl2024effectiveness} opinion distributions were obtained by directly prompting GPT-3.5-turbo to generate them. However, since LLMs are designed to predict the next tokens in a sequence, extracting opinion distributions from them directly is inherently limited \citep{pavl2024effectiveness}.
In 
\citep{sorensen2024roadmappluralisticalignment} log probabilities were sampled and averaged across runs to derive distributions, while in \citep{santurkar2023whose} the log probabilities of plausible tokens were taken for each answer and normalised to extract opinion distributions from two model families. A similar approach was adopted by \citep{lee2023can} using log probabilities and additionally 
repeated prompting {(\small referred to as Monte Carlo estimation)} %
to derive the opinion distributions from four different model families. So far, these simpler approaches have shown limited alignment.

\section{Methodology}
\vspace{-0.2cm}
\todo[]{add motivation for using various temps.... label variation...}
Information on temperature parameter settings was not provided by \citep{lee2023can},  \citep{santurkar2023whose} or \citep{sorensen2024roadmappluralisticalignment}. 
Thus, we 
implement both approaches {\small (Monte Carlo \& Log Probability estimation)}  with the maximum (2.0), the minimum ($\sim$0) and a medium (0.8) temperature setting and evaluate the outputs against the direct approach in \citep{pavl2024effectiveness} on subjective classification tasks.
For the model we employ OpenAI's \texttt{gpt-3.5-turbo} model. 
Across all runs, the top-p parameter is maintained at 1.0 
and all sampling is performed with a repetition of $M=10$  iterations. Due to API rate limitations, we conduct the experiments on a representative subset of 101 examples from each dataset's test set
as provided by \citep{leonardelli2023semeval}.

\textbf{Notation: } \,
We have data $\left\{\left(x_n , y_n\right)\right\}_{n=1}^N$  where for each pair, $x_n$ is a text instance and $y_n \in \triangle^{C-1}$ is a probability distribution. Here, $C$ is the number of categories, and $\triangle^{C-1}$ denotes the probability simplex with dimension $C-1$.
These distributions are obtained by aggregating annotator votes to obtain $\mathbb{P}_{ann}$ and when careful consideration is given to the number, diversity and quality of annotators; we assume that the resulting label distribution approximates the true conditional distribution $\mathbb{P}_{ann} \sim \mathbb{P}$. We use $\hat{y}_n$ to denote a probability distribution obtained from an LLM through the methods described in section \ref{subsection:sampling_logits}.



\vspace{-0.1cm}
\subsection{Methods}
\vspace{-0.2cm}
\label{subsection:sampling_logits}
\textbf{Direct Estimation:} This approach simply involves prompting the LLM to generate an opinion distribution. \vspace{1mm} \\
\textbf{Monte Carlo Estimation (MCE):} We adopt the approach and terminology of \citep{lee2023can} and sample repeated responses $r_m$ (where $m \in [1..M]$) from a model considering only valid options $v_c$ for a given class
$c$ with which we generate the LLM opinion distribution as follows:\footnote{\textbf{1} simply denotes an indicator function}
\begin{equation}\label{eq:sampling}
    \mathbf{P}_{\hat{y}}(c | x) \approx \frac{1}{M*} \sum_{m=1}^{M} \mathbf{1}_{r_{m} \in v_c} \;,
    \vspace{-3mm}
\end{equation} 

where $M* =\sum_{m=1}^{M}  \mathbf{1}_{\exists \, j \in [1..C]\; r_m \in v_j }$, represents the number of iterations excluding invalid responses. \vspace{1mm} \\
\textbf{Log Probability Estimation (LPE):} Similar to the approach in \citep{lee2023can} we also extract the log probabilities ($lp$) of top-k token candidates\footnote{k is set to 10} to estimate the categorical distribution of the labels. 
However, in contrast to using a single instance, we repeatedly sample the log probabilities due to their values varying across iterations. 
Subsequently, we compute the average over the samples to form the opinion distribution for a given $x_n$: 
\begin{equation}\label{eq:logits}
        \mathbf{P}_{\hat{y}}(c | x) \approx  \frac{1}{M*} \sum_{m=1}^{M} \frac{ \sum_{i=1}^{k} e^{\text{lp}_{im}} \cdot \mathbf{1}_{r_{m}\in v_c} }{ \sum_{j=1}^{C} \sum_{i=1}^{k} e^{\text{lp}_{im}} \cdot \mathbf{1}_{r_m\in v_j}} 
    \vspace{-2mm}
\end{equation}

\subsection{Evaluation metrics}
\vspace{-0.2cm}
Given the absence of a definitive standard for evaluating distributional alignment, we assess performance using a variety of metrics. First, 
we use widely adopted metrics, such as cross-entropy (CE) and Jensen-Shannon divergence (JSD).
Furthermore, we include the Manhattan (L1) distance as it is shown to work well for evaluating human opinion distributions in binary settings \citep{rizzi_2024_soft}.
We also include the \textit{Human Distribution Calibration Error} (DistCE) and \textit{Human Entropy Calibration Error} (EntCE)
proposed in \citep{baan2022stop}.
DistCE simply uses the total variation distance (TVD) between the two distributions $\text{DistCE}(y_n,\hat{y}_n)= \text{TVD}({y_n},{\hat{y}_n})$, reflecting how much they diverge from one another, while EntCE aims to capture the agreement between the model's uncertainty $\text{H}(\hat{y}_n)$ and the human disagreement $\text{H}(y_n)$ for a given instance $x_n$: $\text{EntCE}(y_n,\hat{y}_n)= \text{H}(\hat{y}_n) - \text{H}(y_n)$ \citep{baan2022stop}. $H(.)$ corresponds to entropy.
Given that the selected evaluation metrics are instance-level metrics $x_n$, we aggregate them by computing the average. For the EntCE metric, this involves first taking the absolute value: $\mathbb{E}[|\text{EntCE}|]$.

\subsection{Datasets}
\vspace{-0.25cm}
We evaluate the {different} methods on the subjective binary classification tasks from the SemEval 2023 shared task on 'Learning with Disagreement'. All datasets are annotated by multiple annotators and address tasks on abusiveness/offensiveness detection.
\vspace{0.5cm}

\begingroup
\setlength{\tabcolsep}{2.5pt} 
\renewcommand{\arraystretch}{0.9} 
\begin{table}[ht]
\vspace{-0.5cm}
\caption{\small Dataset statistics \citep{leonardelli2023semeval}  - ({language code: ISO 639})} 
\vspace{2mm}
\centering
\begin{tabular}[t]{llccccc} 
\toprule
\arraycolsep=0.9pt\def\arraystretch{0.8}
$\begin{array}{l}
\textbf{\footnotesize Dataset} \end{array}$ 
& 
\arraycolsep=0.9pt\def\arraystretch{0.8}
$\begin{array}{c}
\textbf{{\footnotesize Task}}
\end{array}$
& 
\arraycolsep=0.9pt\def\arraystretch{0.8}
$\begin{array}{c}
\textbf{{\footnotesize Language}}
\end{array}$
&
\arraycolsep=0.9pt\def\arraystretch{0.8}
$\begin{array}{c}
\textbf{{\footnotesize \# annotators}}
\end{array}$
&
\arraycolsep=0.9pt\def\arraystretch{0.5}
$\begin{array}{c}
     \text{\textbf{\footnotesize \# items}}
\end{array}$  &
\arraycolsep=0.9pt\def\arraystretch{0.8}
$\begin{array}{c}
    \text{\textbf{\footnotesize \% full annotator agreement}} 
\end{array}$ \\
\midrule
{\footnotesize HS-Brexit} & 
    \arraycolsep=0.1pt\def\arraystretch{0.5} 
    \begin{tabular}[c]{@{}l@{}}
        \footnotesize {Offensiveness detection} 
    \end{tabular} &
{\footnotesize en} &
\footnotesize 6 &
\footnotesize 1120
& {\footnotesize 69\%} \\
{\footnotesize ConvAbuse} &
\arraycolsep=0.1pt\def\arraystretch{0.5} 
    \begin{tabular}[c]{@{}l@{}}
        \footnotesize {Abusiveness detection} 
    \end{tabular}  &
{\footnotesize en} &
\footnotesize 2-8 &
\footnotesize 4050 
& {\footnotesize 86\%} \\

{\footnotesize MD-Agree.} &
    \arraycolsep=0.1pt\def\arraystretch{0.5} 
    \begin{tabular}[c]{@{}l@{}}
        \footnotesize {Offensiveness detection} 
    \end{tabular} &
{\footnotesize en} &
\footnotesize 5 &
\footnotesize 10753
& {\footnotesize 42\%} \\
\bottomrule
\end{tabular}
\label{table:datasets}
\end{table}
\endgroup

\vspace{-0.35cm}
\section{Results}
\vspace{-0.2cm}
Table \ref{table:eval_metrics} presents the results for all metrics across the datasets. 
MCE consistently achieves the lowest scores for each dataset, though the differences between MCE and LPE are marginal.
To gain a more detailed understanding of the results, we visualise the entropy of GPT compared to that of humans in Figure \ref{fig:entropy_histograms}. This figure includes the direct method and contrasts it with all MCE runs. The MCE approach shows a stronger alignment between model and human entropy compared to the direct prompting approach. Consistent with expectations, we also observe that as the temperature decreases, the model shows a greater consistency in its sampled outputs. This results in a higher frequency of zero-entropy distributions, which diverge more from the human entropy, see Figure \ref{fig:hsb_sampl_2}-\ref{fig:hsb_sampl_0}, \ref{fig:ca_sampl_2}-\ref{fig:ca_sampl_0},  
\ref{fig:mda_sampl_2}-\ref{fig:mda_sampl_0}.

In addition to the histogram plots, we also visualise the relationship between GPT and human entropy levels, as in \citep{lee2023can}, for the best performing MCE run for MDA, see Figure 
\ref{fig:scatter_mda}. For 
other 
datasets see Appendix \ref{appendix:supporting_figures}. 
The scatter-plot displays each instance with its human and LLM entropy, with point size showing their count. 

\vspace{-2mm}
\begin{table}[h]
    \begin{minipage}[t]{0.5\textwidth}
        \setlength{\tabcolsep}{0.15pt}
        \renewcommand{\arraystretch}{1.11}
        \caption{\small Results on subsets of the SemEval2023 classification datasets with different estimation methods: \textbf{MC} refers to 'monte carlo estimation', \textbf{LPE} to the 'log probability estimation' and \textbf{direct} refers to prompting for opinion distributions }
        \label{table:eval_metrics}
        \vspace{3.6mm}   
        \centering
        \begin{tabular}{lrccccccc}
        \toprule
        {} &
        {} &
        \textbf{\footnotesize Direct} &
        \arraycolsep=.9pt\def\arraystretch{0.3}
        $\begin{array}{c}
             \text{\textbf{\footnotesize MC}} \\     
             \text{{\scriptsize T=2}}
        \end{array}$ 
        \hfill
        &
        \arraycolsep=.9pt\def\arraystretch{0.3}
        $\begin{array}{c}
             \text{\textbf{\footnotesize LP}} \\     
             \text{{\scriptsize T=2}}
        \end{array}$ 
        \hfill
        &
        \arraycolsep=.9pt\def\arraystretch{0.3}
        $\begin{array}{c}
             \text{\textbf{\footnotesize MC}} \\     
             \text{{\scriptsize T=0.8}}
        \end{array}$ 
        \hfill
        &
        \arraycolsep=.9pt\def\arraystretch{0.3}
        $\begin{array}{c}
             \text{\textbf{\footnotesize LP}} \\     
             \text{{\scriptsize T=0.8}}
        \end{array}$ 
        &
        \arraycolsep=.9pt\def\arraystretch{0.3}
        $\begin{array}{c}
             \text{\textbf{\footnotesize MC}} \\     
             \text{{\scriptsize T$\approx$0}}
        \end{array}$ 
        &
        \arraycolsep=.9pt\def\arraystretch{0.3}
        $\begin{array}{c}
             \text{\textbf{\footnotesize LP}} \\     
             \text{{\scriptsize T$\approx$0}}
        \end{array}$ \\
        \toprule
        \multirow{5}{*}{\rotatebox[origin=c]{90}{\footnotesize \textbf{HS Brexit}}} &
        {\footnotesize $CE \downarrow$} & \footnotesize {4.65} & \footnotesize{0.52} & \footnotesize{0.42} & \footnotesize{0.34} & \footnotesize{0.42} & \footnotesize\textbf{0.25} & \footnotesize{0.40}  
        \\
        & {\footnotesize $JSD \downarrow$} & \footnotesize {0.14} & \footnotesize \textbf{0.04} & \footnotesize {0.05} & \footnotesize {0.05} & \footnotesize {0.05} & \footnotesize {0.05} & \footnotesize {0.05}\\
        & {\footnotesize $EntCE \downarrow$} & \footnotesize {0.37} & \footnotesize \textbf{0.18} & \footnotesize{0.20} & \footnotesize{0.21} & \footnotesize{0.20} & \footnotesize{0.21} & \footnotesize{0.19} \\
        & {\footnotesize $DistCE \downarrow$} & \footnotesize {0.36} & \footnotesize \textbf{0.12} & \footnotesize {0.13} & \footnotesize{0.13} & \footnotesize{0.13} & \footnotesize{0.13} & \footnotesize{0.13} \\
        & {\footnotesize $L1 \downarrow$} & \footnotesize{0.72} & \footnotesize \textbf{0.25} & \footnotesize {0.26} & \footnotesize{0.27} & \footnotesize{0.26} & \footnotesize{0.26} & \footnotesize{0.26} \\
        
        \midrule
        \multirow{5}{*}{\rotatebox[origin=c]{90}{\footnotesize \textbf{ConvAbuse}}} &
        {\footnotesize $CE \downarrow$} & \footnotesize 3.45 & \footnotesize 1.25 & \footnotesize 0.72 & \footnotesize \textbf{0.68} & \footnotesize 0.72 & \footnotesize 0.76 & \footnotesize 0.72  
        \\
        & {\footnotesize $JSD \downarrow$} & \footnotesize 0.07 & \footnotesize \textbf{0.04} & \footnotesize \textbf{0.04} & \footnotesize \textbf{0.04}  & \footnotesize \textbf{0.04}  & \footnotesize \textbf{0.04} & \footnotesize \textbf{0.04} \\
        & {\footnotesize $EntCE \downarrow$}  & \footnotesize 0.45 & \footnotesize 0.16 & \footnotesize 0.10 & \footnotesize \textbf{0.08}  & \footnotesize 0.11  & \footnotesize \textbf{0.08} & \footnotesize 0.10  \\
        & {\footnotesize $DistCE \downarrow$} & \footnotesize 0.20 & \footnotesize 0.10 & \footnotesize 0.08 & \footnotesize \textbf{0.07}  & \footnotesize 0.08  & \footnotesize 0.08 & \footnotesize 0.08 \\
        & {\footnotesize $L1 \downarrow$} & \footnotesize 0.40 &  \footnotesize 0.20 & \footnotesize 0.16 & \footnotesize \textbf{0.15}  & \footnotesize 0.16  & \footnotesize \textbf{0.15} & \footnotesize 0.16  \\
        
        \midrule
        \multirow{5}{*}{\rotatebox[origin=c]{90}{\footnotesize \textbf{MD-Agree.}}} &
        {\footnotesize $CE \downarrow$} & \footnotesize 3.89 & \footnotesize 1.35 & \footnotesize 1.28 & \footnotesize \textbf{1.19} & \footnotesize 1.25 & \footnotesize 1.26 & \footnotesize 1.25 \\
        & {\footnotesize $JSD \downarrow$}  & \footnotesize 0.14 & \footnotesize \textbf{0.09} & \footnotesize 0.10 & \footnotesize 0.11 & \footnotesize 0.10 & \footnotesize 0.12 & \footnotesize 0.10 
        \\
        & {\footnotesize $EntCE \downarrow$}  & \footnotesize 0.27 & \footnotesize \textbf{0.24} & \footnotesize 0.26 & \footnotesize 0.30 & \footnotesize 0.27 & \footnotesize 0.31 & \footnotesize 0.26 
        \\
        & {\footnotesize $DistCE \downarrow$} & \footnotesize 0.36 & \footnotesize \textbf{0.22} & \footnotesize 0.24 & \footnotesize 0.25 & \footnotesize 0.24 & \footnotesize 0.26 & \footnotesize 0.24  
        \\
        & {\footnotesize $L1 \downarrow$} & \footnotesize 0.71 & \footnotesize \textbf{0.44} & \footnotesize 0.49 & \footnotesize 0.49 & \footnotesize 0.48 & \footnotesize 0.51 & \footnotesize 0.48 \\\bottomrule 
        \end{tabular}
        
    \end{minipage}
    \hfill
    \begin{minipage}[t]{0.45\textwidth}
        \begin{minipage}[t]{\textwidth}
        \setlength{\tabcolsep}{7pt}
        \centering
        \renewcommand{\arraystretch}{0.5}
        \caption{\small ECE results on subsets of the SemEval2023 classification datasets with MC estimation methods. Lower is better.}
        \label{table:ece_metrics}
        \vspace{1mm} 
        \begin{tabular}{lccc}
            \toprule
            {} &
            \arraycolsep=.9pt\def\arraystretch{0.3}
            $\begin{array}{c}
                 \text{\textbf{\footnotesize MC}} \\     
                 \text{{\scriptsize T=2}}
            \end{array}$ 
            \hfill
            &
            \arraycolsep=.9pt\def\arraystretch{0.3}
            $\begin{array}{c}
                 \text{\textbf{\footnotesize MC}} \\     
                 \text{{\scriptsize T=0.8}}
            \end{array}$ 
            \hfill
            &
            \arraycolsep=.9pt\def\arraystretch{0.3}
            $\begin{array}{c}
                 \text{\textbf{\footnotesize MC}} \\     
                 \text{{\scriptsize T=1e-06}}
            \end{array}$ 
            \\ 
            \toprule
            {\footnotesize HS-Brexit} & \footnotesize \textbf{0.948} & \footnotesize {0.977} & \footnotesize {0.983} 
            \\
            {\footnotesize ConvAbuse} & \footnotesize \textbf{0.951} & \footnotesize {0.99} & \footnotesize{0.99} 
            \\
            {\footnotesize MD-Agree.} & \footnotesize \textbf{0.922} & \footnotesize {0.971} & \footnotesize {0.99} 
            \\
            \bottomrule 
            \end{tabular}
        \end{minipage}
        \begin{minipage}[b]{\textwidth}
        \vspace{3mm} 
            \centering
            \includegraphics[width=\textwidth, height=5.55cm]{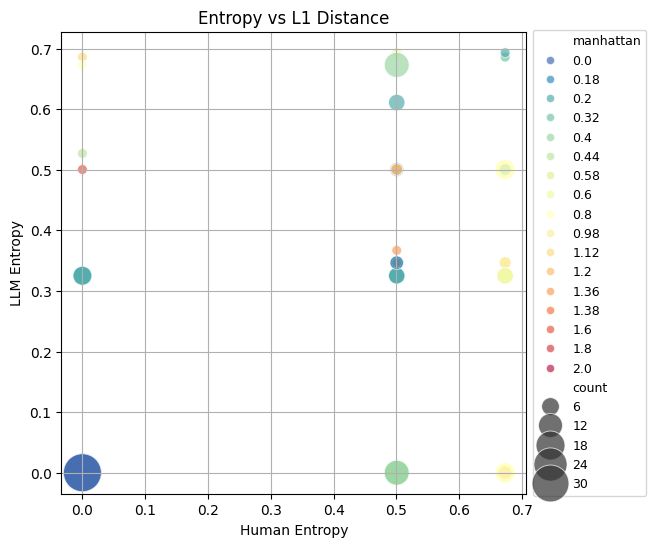} 
            \captionof{figure}{\small MD-Agree. T=2.0; ideally, if entropies align, a model should display a diagonal line from bottom left to top right;
            lower L1 is better}
            \label{fig:scatter_mda}
        \end{minipage}
    \end{minipage}
\end{table}

\vspace{-0.4cm}
\section{Discussion}
\vspace{-0.2cm}
The results indicate that sampling produced more closely aligned distributions than direct extraction of opinion distributions, primarily due to improved alignment in cases with full label agreement among humans (0 entropy), as visible in Figure \ref{fig:entropy_histograms} and additional supporting figures in Appendix \ref{appendix:supporting_figures}.
In both \citep{lee2023can} and \citep{santurkar2023whose}, the authors observed that the GPT-based model (\texttt{text-davinci-003}) generated overly confident opinion distributions (near 0 entropy levels).
In contrast to their work, we explored various temperature settings with \texttt{gpt-3.5-turbo} and find that sampling with higher temperatures reduced such "overconfident" predictions, resulting in distributions that are more closely calibrated with human opinions. 

Figure \ref{fig:scatter_mda} and Appendix \ref{appendix:supporting_figures}, show that points along the diagonal generally exhibit low L1 distances, while those further from the diagonal have higher values, 
suggesting that L1 may capture aspects of binary opinion alignment. 
We also see a few samples where the model entropy equals the human entropy (0.5) but the L1 distance indicates the probabilities are far from aligning. This is due to entropy $\text{H}(X) := -\sum_{x \in \mathcal{X}} p(x) \log p(x)$ being invariant to permutations of the the probability values. Thus, it should be noted that entropy, despite providing valuable insights for evaluation \citep{lee2023can, santurkar2023whose, pavl2024effectiveness}, can be misleading when considered in isolation for human distributional alignment.


We interpret the distributional outputs as indicative of opinions; however, they are also commonly viewed as reflecting model uncertainty. These two differing perspectives are highlighted in \citep{baan_2024_interpreting}, emphasising that both model confidence and human label variation are essential for developing trustworthy and fair NLP systems.
Results in \citep{gao2024spuq} indicated that sampling with an increase in temperature improves model uncertainty calibration. We, thus additionally evaluated the MCE runs with the 
expected calibration error (ECE), which measures model certainty calibration, and found preliminary results to be consistent with \citep{gao2024spuq}, see Table \ref{table:ece_metrics}.
Further research is needed to understand the relationship between  the variance in human opinions and model uncertainty. 


\begin{figure*}[ht!]
  \centering
  \begin{subfigure}[t]{0.24\textwidth}
    \centering
    \includegraphics[width=\textwidth]{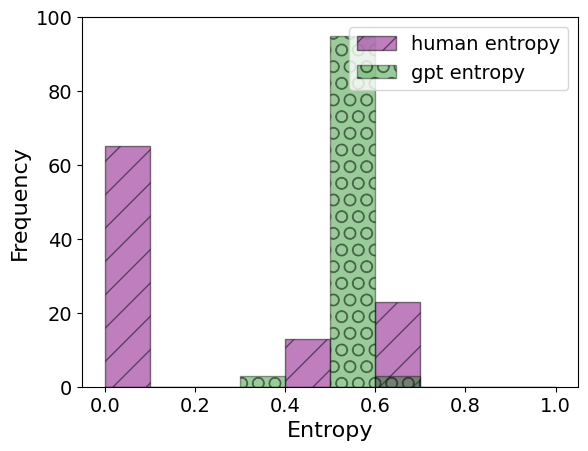}
    \caption{HSB - direct}
    \label{fig:hsb_direct}
  \end{subfigure}
  \hfill
  \begin{subfigure}[t]{0.24\textwidth}
    \centering
    \includegraphics[width=\textwidth]{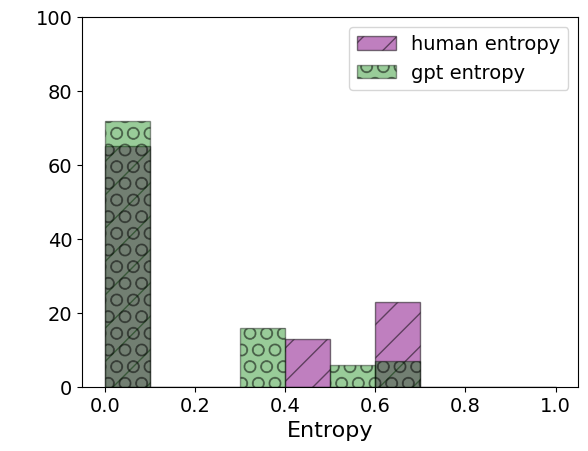}
    \caption{HSB - MC, T=2}
    \label{fig:hsb_sampl_2}
  \end{subfigure}
  \hfill
  \begin{subfigure}[t]{0.24\textwidth}
    \centering
    \includegraphics[width=\textwidth]{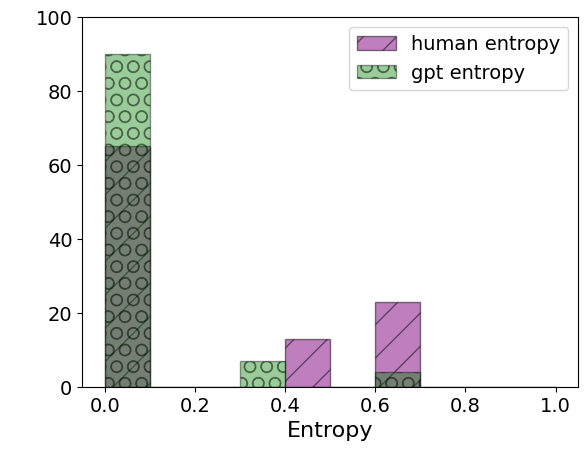}
    \caption{HSB - MC, T=0.8}
    \label{fig:hsb_sampl_08}
  \end{subfigure}
  \hfill
  \begin{subfigure}[t]{0.24\textwidth}
    \centering
    \includegraphics[width=\textwidth]{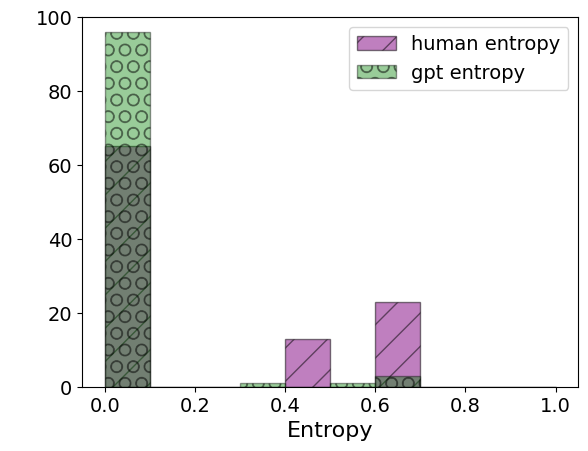}
    \caption{HSB - MC, T$\approx$0}
    \label{fig:hsb_sampl_0}
  \end{subfigure}


  \centering
  \begin{subfigure}[t]{0.24\textwidth}
    \centering
    \includegraphics[width=\textwidth]{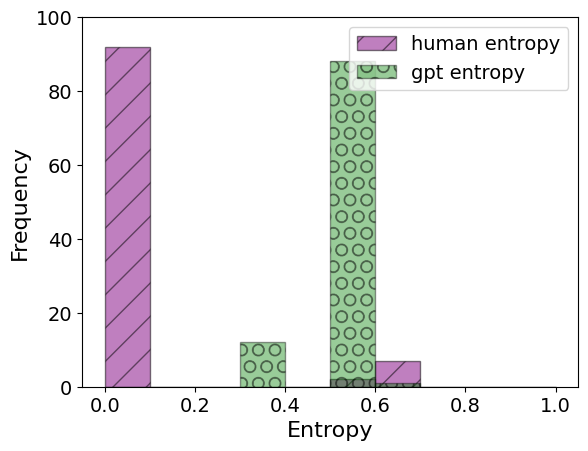}
    \caption{CA - direct}
    \label{fig:ca_direct}
  \end{subfigure}
  \hfill
  \begin{subfigure}[t]{0.24\textwidth}
    \centering
    \includegraphics[width=\textwidth]{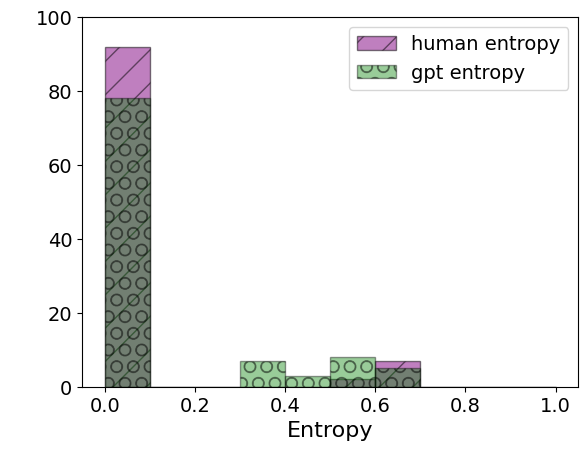}
    \caption{CA - MC, T=2}
    \label{fig:ca_sampl_2}
  \end{subfigure}
  \hfill
  \begin{subfigure}[t]{0.24\textwidth}
    \centering
    \includegraphics[width=\textwidth]{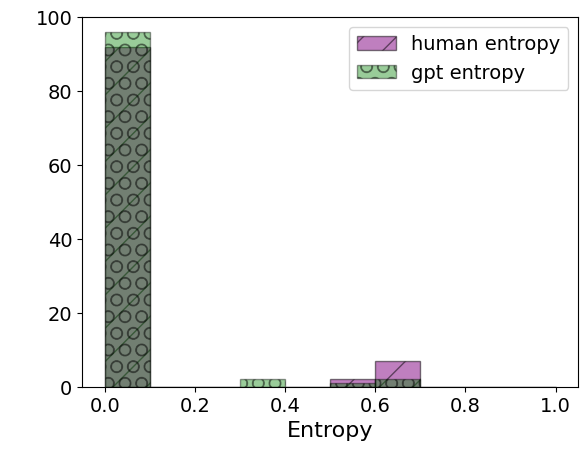}
    \caption{CA - MC, T=0.8}
    \label{fig:ca_sampl_08}
  \end{subfigure}
  \hfill
  \begin{subfigure}[t]{0.24\textwidth}
    \centering
    \includegraphics[width=\textwidth]{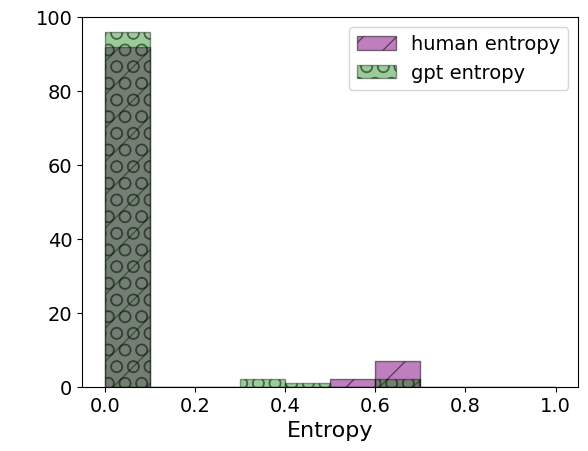}
    \caption{CA - MC, T$\approx$0}
    \label{fig:ca_sampl_0}
  \end{subfigure}

  \centering
  \begin{subfigure}[t]{0.23\textwidth}
    \centering
    \includegraphics[width=\textwidth]{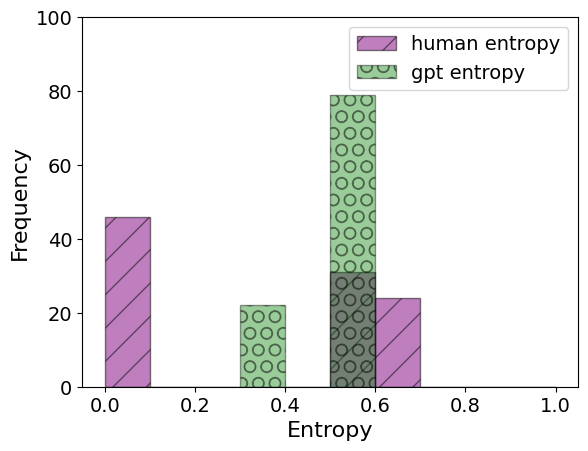}
    \caption{MD-Agr. - direct}
    \label{fig:mda_direct}
  \end{subfigure}
  \hfill
  \begin{subfigure}[t]{0.23\textwidth}
    \centering
    \includegraphics[width=\textwidth]{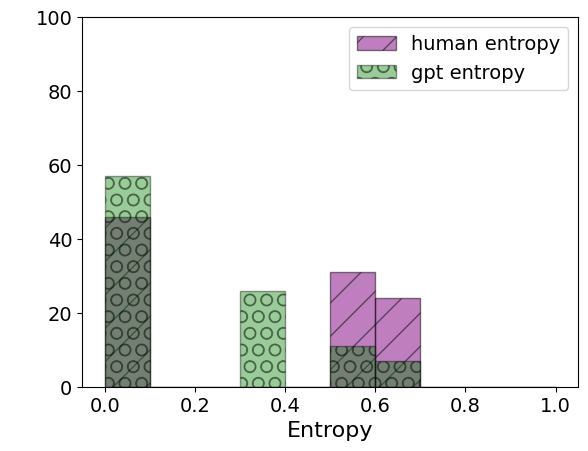}
    \caption{MD-Agr. - MC, T=2}
    \label{fig:mda_sampl_2}
  \end{subfigure}
  \hfill
  \begin{subfigure}[t]{0.23\textwidth}
    \centering
    \includegraphics[width=\textwidth]{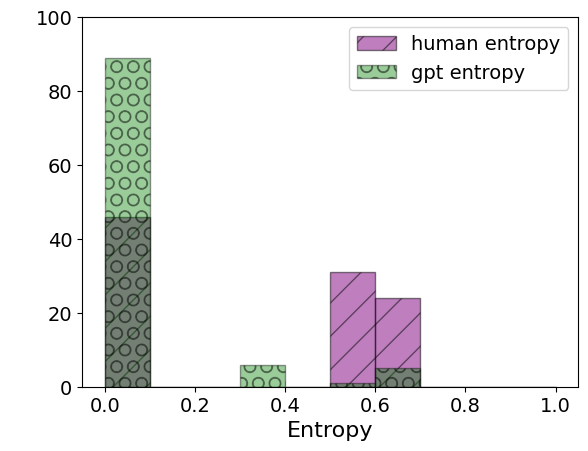}
    \caption{MD-Agr. - MC, T=0.8}
    \label{fig:mda_sampl_08}
  \end{subfigure}
  \hfill
  \begin{subfigure}[t]{0.23\textwidth}
    \centering
    \includegraphics[width=\textwidth]{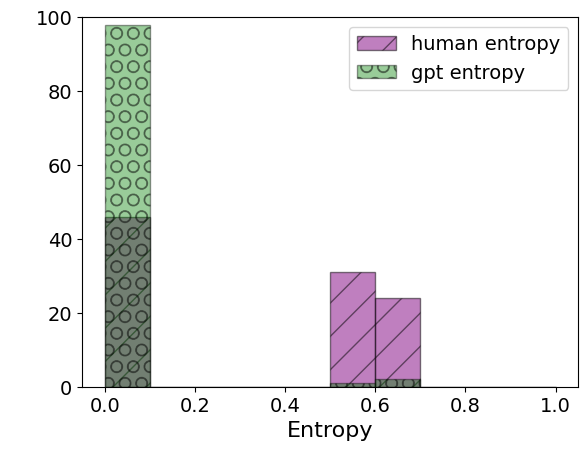}
    \caption{MD-Agr. - MC, T$\approx$0}
    \label{fig:mda_sampl_0}
  \end{subfigure}
  \caption{\small Histograms of human and GPT entropy levels for SemEval2023 datasets: directly generated GPT distributions vs. those generated with the MCE method.}
  \label{fig:entropy_histograms}
  \vspace{-0.17cm}
\end{figure*}

\vfill
\section{Conclusion}
\vspace{-0.2cm}
We assessed three techniques for deriving opinion distributions from GPT in this empirical study and find that response sampling {\small(MC)} with higher temperatures produced distributions more closely reflecting human distributions of abusiveness and offensiveness.
We highlight the potential role of temperature in improving opinion alignment, and emphasise the importance of reporting on such parameter settings in studies, particularly in contexts of potential model overconfidence. 
We encourage further research on its generalisation to other tasks and suggest investigating how social and cultural value disagreement affects overall model uncertainty.  Accurate opinion modeling could influence decision-making, requiring further research into 
potential biases, ethical implications, and broader effects on society.

\vfill
\section{Limitations}
\vspace{-0.2cm}
A limitation of our approach is the assumption that $\mathbb{P}_{ann} \sim \mathbb{P}$, 
especially given the number of annotators per instance. The data in ConvAbuse, for instance, was predominantly annotated by two to three annotators, whereas the other two datasets have at least five annotations per sample. 
To achieve a more precise approximation of the true conditional distribution, an increased number of high-quality annotators per instance would be preferable. Furthermore, due to rate limitations, the sample size was constrained but carefully chosen to represent the overall distribution of the full dataset. 

\bibliographystyle{unsrt}
\bibliography{neurips_2024}

\appendix
\section{Prompts}
\vspace{-0.2cm}
\label{sec:prompts}
\paragraph{Hate-Speech Brexit:} "You are an expert annotator in "hate speech detection", who considers multiple worldviews and cultural nuances. Is the text, which is delimited by three dashes (-), "hate speech"? Respond only with 'yes' or 'no'. Only respond with one word!" 
\paragraph{Conversational Abuse:} "You are an expert annotator in "abusiveness detection", who considers multiple worldviews and cultural nuances. Is the text, which is delimited by three dashes (-), "abusive"? Respond only with 'yes' or 'no'. Only respond with one word!"
\paragraph{Multi-Domain Agreement:} "You are an expert annotator in "offensiveness detection", who considers multiple worldviews and cultural nuances. Is the text, which is delimited by three dashes (-), "offensive"? Respond only with 'yes' or 'no'. Only respond with one word!"

\section{Datasets}
\vspace{-0.2cm}
\label{appendix-datasets}
The datasets used in this study 
have been shared openly online by the authors \citep{Akhtar_Basile_Patti_2021, Leonardelli_Menini_Palmero_Aprosio_Guerini_Tonelli_2021, curry2021convabuse, Almanea_Poesio_2022} for research purposes as part of the shared task on learning from disagreement (SemEval2023) co-located with NAACL. Their website: \href{https://le-wi-di.github.io/}{https://le-wi-di.github.io/} The datasets can be accessed at: \href{https://github.com/Le-Wi-Di/le-wi-di.github.io/blob/main/data\_post-competition.zi}{https://github.com/Le-Wi-Di/le-wi-di.github.io/blob/main/data\_post-competition.zip}

\newpage

\section{Supporting Figures}
\vspace{-0.2cm}
\label{appendix:supporting_figures}

\begin{figure*}[ht!]
  \centering
  \begin{subfigure}[t]{0.47\textwidth}
    \centering
    \includegraphics[width=\textwidth]{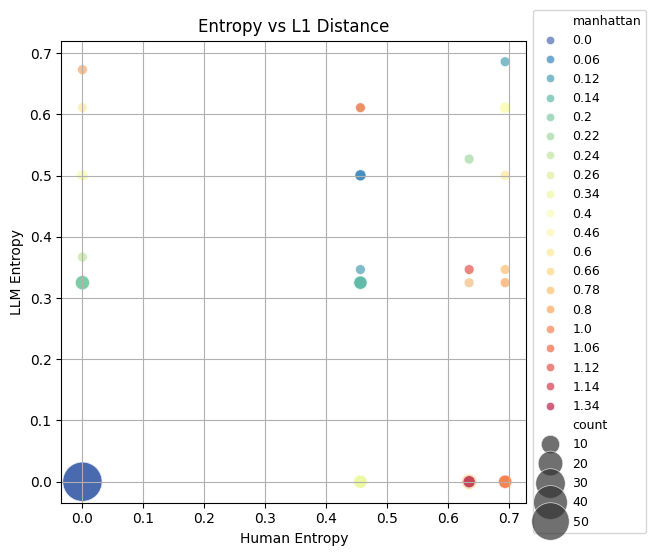}
    \caption{HS-Brexit  - MC, T=2}
    \label{fig:scatter_hsb}
    \end{subfigure}
  \hfill
  \begin{subfigure}[t]{0.47\textwidth}
    \centering
        \includegraphics[width=\textwidth]{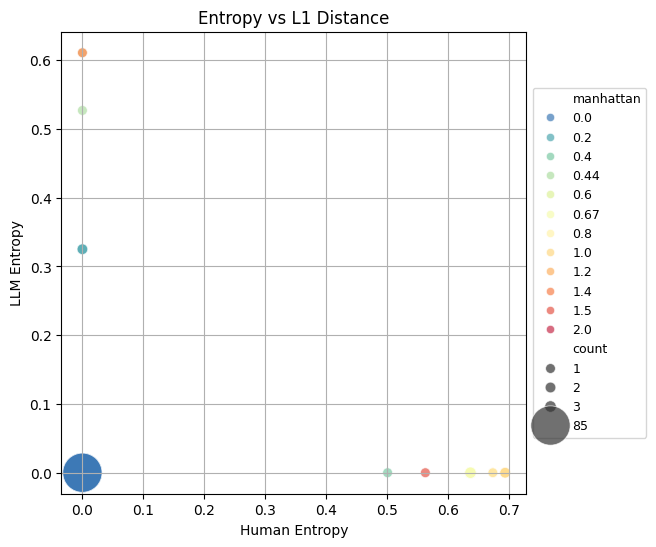}
        \caption{ConvAbuse - MC, T=0.8}
    \label{fig:scatter_ca}
  \end{subfigure}
\caption{ Ideally, if entropies align, a model should display a diagonal line from bottom left to top right; A lower manhattan distance (L1) is better. The majority of samples fall under human agreement {\textit{ (0 entropy)}} with GPT’s prediction confidence high {\textit{ (0 entropy)}} on such samples.
}
\label{fig:all_scatter}
\end{figure*}

\begin{figure*}[ht!]
  \centering
  \begin{subfigure}[t]{0.322\textwidth}
    \centering
        \includegraphics[width=\textwidth]{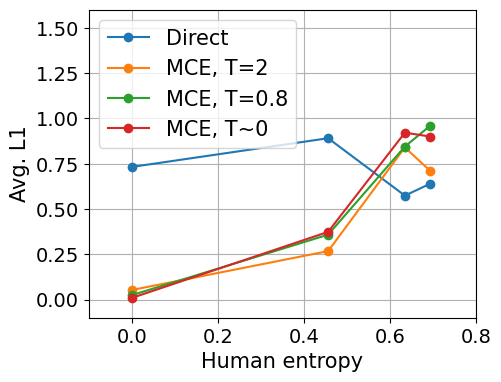}
    \caption{HS-Brexit}
    \label{fig:l1_hsb}
    \end{subfigure}
  \hfill
  \begin{subfigure}[t]{0.32\textwidth}
    \centering
        \includegraphics[width=\textwidth]{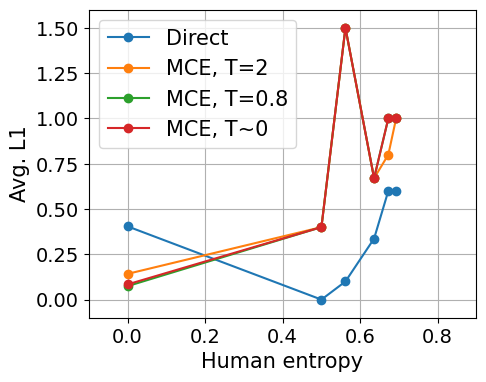}
    \caption{ConvAbuse}
    \label{fig:l1_ca}
  \end{subfigure}
  \hfill
  \begin{subfigure}[t]{0.32\textwidth}
    \centering
        \includegraphics[width=\textwidth]{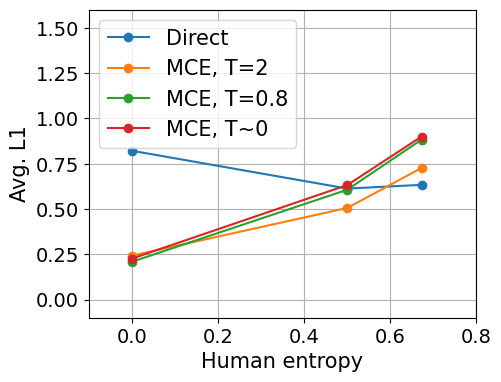}
    \caption{MD-Agree.}
    \label{fig:l1_mda}
  \end{subfigure}
\caption{Average L1-distance \textit{manhattan-distance} from model distributions in relation to entropy values of human opinion distributions:\, The sampling approach (MCE) performs more in line with expectations across all datasets, showing that this method exhibits greater confidence on samples with full human agreement (0 entropy); while the direct method doesn't capture this confidence on easier instances as effectively.}
\label{fig:all}
\end{figure*}

\begin{figure*}[ht!]
  \centering
  \begin{subfigure}[t]{0.95\textwidth}
    \centering
        \includegraphics[width=\textwidth]{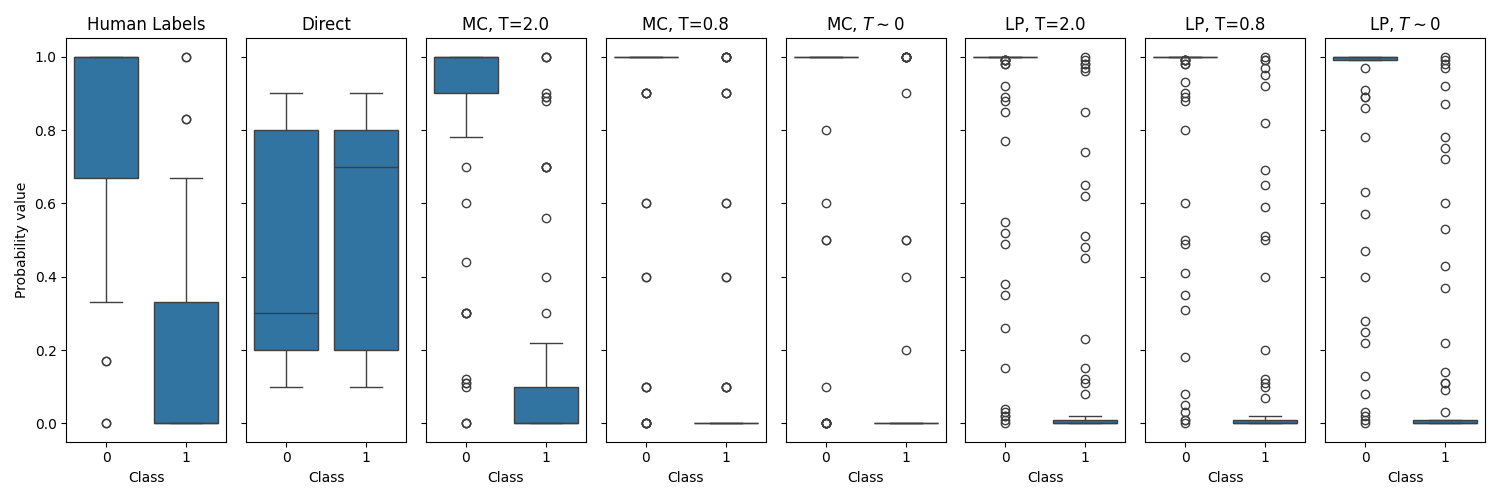}
    \caption{HS-Brexit}
    \label{fig:boxplot_full_hsb}
    \end{subfigure}
  \vfill
  \begin{subfigure}[t]{0.95\textwidth}
    \centering
        \includegraphics[width=\textwidth]{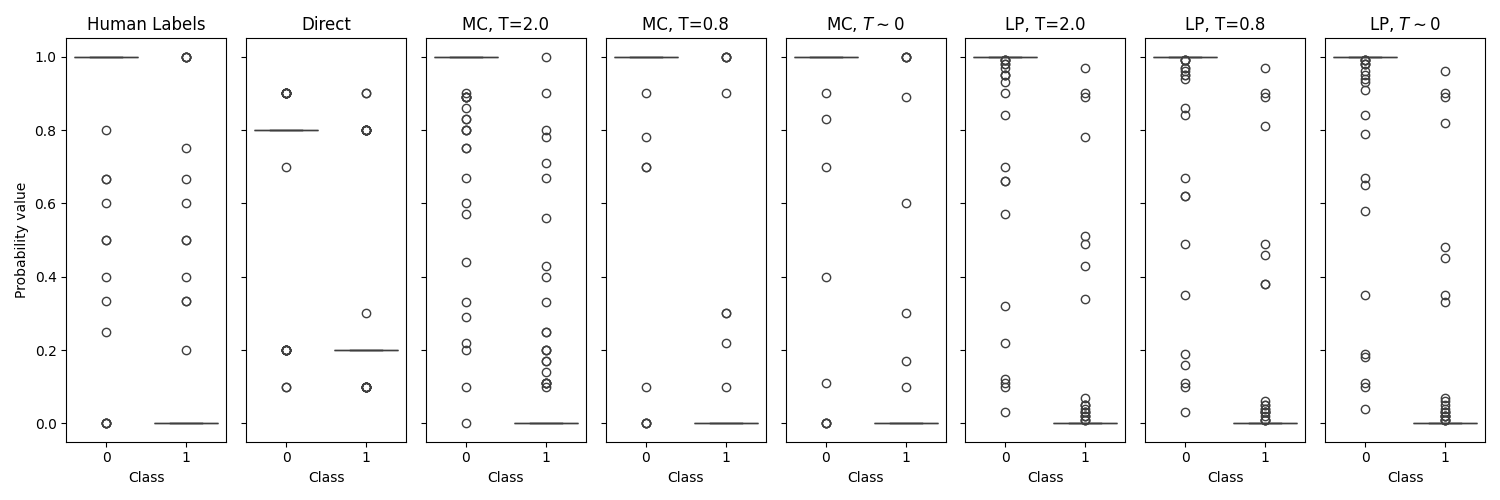}
    \caption{ConvAbuse}
    \label{fig:boxplot_full_ca}
  \end{subfigure}
  \vfill
  \begin{subfigure}[t]{0.95\textwidth}
    \centering
        \includegraphics[width=\textwidth]{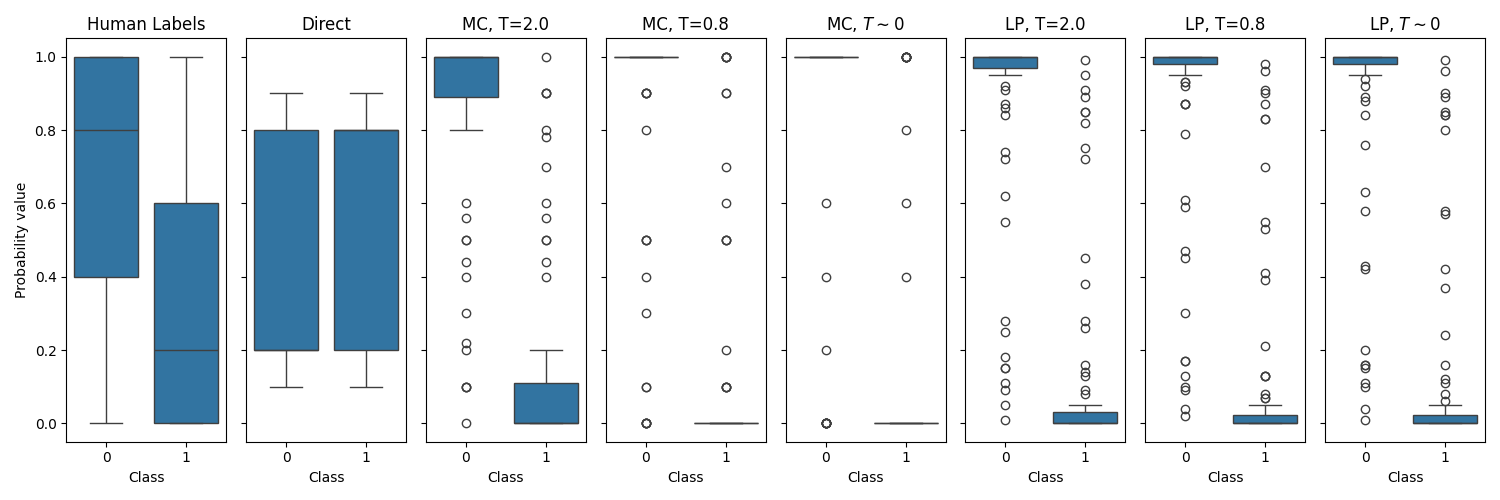}
    \caption{MD-Agree.}
    \label{fig:boxplot_full_mda}
  \end{subfigure}
\caption{Distribution for each run (direct, MC and LP). Human Labels have more variation in HS-Brexit and MD-Agreement datasets.}
\label{fig:boxplots}
\end{figure*}

\end{document}